# Quaternion-Based Robust PCA for Efficient Moving Target Detection and Background Recovery in Color Videos

Liyang Wang, Shiqian Wu, *Senior Member*, *IEEE*, Shun Fang, Qile Zhu, Jiaxin Wu, and Sos Again, *Fellow*, *IEEE*

*Abstract*—Moving target detection is a challenging computer vision task aimed at generating accurate segmentation maps in diverse in-the-wild color videos captured by static cameras. If backgrounds and targets can be simultaneously extracted and recombined, such synthetic data can significantly enrich annotated in-the-wild datasets and enhance the generalization ability of deep models. Quaternion-based RPCA (QRPCA) is a promising unsupervised paradigm for color image processing. However, in color video processing, Quaternion Singular Value Decomposition (QSVD) incurs high computational costs, and rank-1 quaternion matrix fails to yield rank-1 color channels. In this paper, we reduce the computational complexity of QSVD to $o(1)$ by utilizing a quaternion Riemannian manifold. Furthermor, we propose the universal QRPCA (uQRPCA) framework, which achieves a balance in simultaneously segmenting targets and recovering backgrounds from color videos. Moreover, we expand to uQRPCA+ by introducing the Color Rank-1 Batch (CR1B) method to further process and obtain the ideal low-rank background across color channels. Experiments demonstrate our uQRPCA+ achieves State Of The Art (SOTA) performance on moving target detection and background recovery tasks compared to existing open-source methods. Our implementation is publicly available on GitHub at *https://github.com/Ruchtech/uQRPCA*.

## I. INTRODUCTION

MOVING target detection is a common and important problem in computer vision and has been widely applied in security [1], object tracking [2], and human activity recognition [3]. Although static cameras can capture diverse real-world color video datasets for this task, it remains challenging to train deep learning models due to the lack of accurate target annotations, which limits model generalization ability. To mitigate this, existing approaches alter target backgrounds in existing datasets, thereby diversifying the data and improving reconstruction [4] or perception [5]. In addition, [6] proposes a model based on Generative Adversarial Networks that automatically generates both a target segmentation mask and a corresponding background from a single image, which are then combined through permutation to produce plausible composite images. This raises a critical question: ***Can we achieve efficient and accurate moving target detection while simultaneously recovering the background from color video?*** Although there deep learning approaches have demonstrated remarkable performance in video foreground–background separation [7], [8], [9], their applicability remains constrained in various scenarios due to the heavy reliance on large-scale annotated datasets or intensive computational resources.

Robust Principal Component Analysis (RPCA) method [10] has emerged as a popular unsupervised technique for video analysis, with applications in moving target detection [11], background recovery [12], denoising [13], and infrared small target detection [14]. In the first two applications, RPCA [15] typically assumes that a video matrix $D$ can be decomposed into two components: $D = L + S$, where $L$ represents a low-rank background matrix, and $S$ is a sparse matrix capturing moving targets. However, these two tasks are often treated as separate research directions with limited joint studying:

***Moving Target Detection in Grayscale Videos:*** Gao et al. [16] developed a 2-pass approach that leverages Salient Sparse Blocks (SSB) and demonstrated its effectiveness across a wide range of realistic and complex situations. Liu et al. [17] extended this idea by introducing a Structured Sparsity Norm (SSN) to eliminate non-target SSB caused by sudden variations. Fan et al. [18] further enhanced this approach by incorporating the SSN into Total Variation regularized RPCA (TVRPCA) [19] to better suppress noise. Despite these advances, significant limitations remain. The effectiveness of SSN diminishes in scenarios where sudden variations are critical, such as anomaly detection [20]. Moreover, even accurately detected target SSB still contains noise and fails to adaptively balance noise suppression and target detail preservation.

***Background Recovery in Color Videos:*** To broaden practical applications and improve background recovery

This paragraph of the first footnote will contain the date on which you submitted your paper for review, which is populated by IEEE. It is IEEE style to display support information, including sponsor and financial support acknowledgment, here and not in an acknowledgment section at the end of the article.

L. Wang, S. Wu, S. Fang, Q. Zhu, and J. Wu are with the Institute of Robotics and Intelligent Systems, Wuhan University of Science and Technology, Wuhan, China (e-mail: liyangwang520@gmail.com; shiqian.wu@wust.edu.cn; fangshun97@wust.edu.cn; zhuqile@wust.edu.cn; wust_jiaxin.wu@hotmail.com).

S. Agaian was with the Department of Computer Science, College of Staten Island, City University of New York, New York, NY 10314 USA (e-mail: sos.agaian@csi.cuny.edu).

Color versions of one or more of the figures in this article are available online at http://ieeexplore.ieee.org

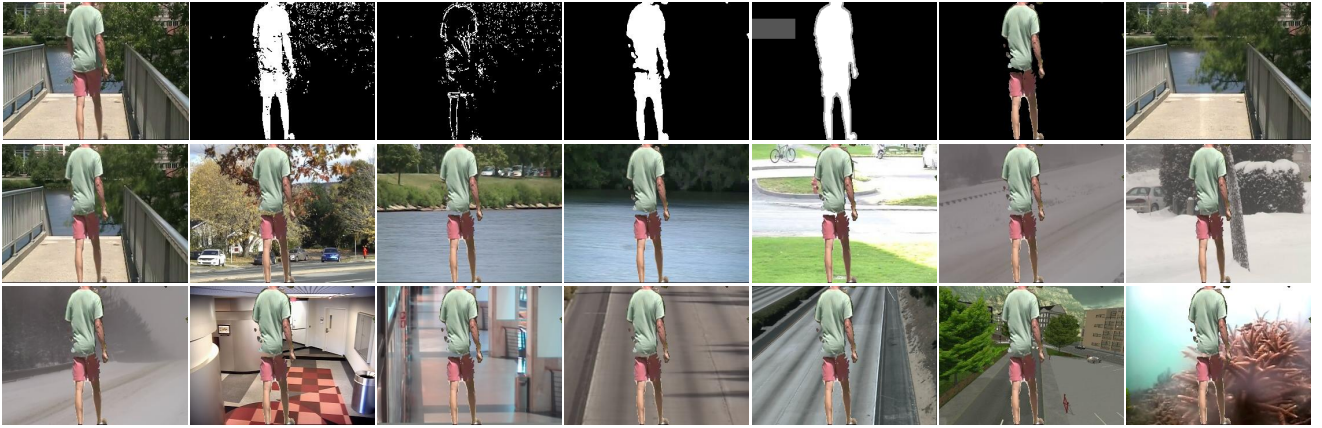

Fig. 1. Visual results on synthetic color video sequences generated by our proposed method, uQRPCA+. The first row (from left to right) shows: original video clip, recovered sparse component, noise component, detected target map, ground-truth map, final detection result, and recovered background. The second and third rows illustrate newly synthesized sequence clips obtained by compositing various recovered backgrounds with final detected target.

performance, Xu et al. [21] proposed a double-weighted RPCA approach based on the estimated rank prior. Fang et al. [22] further introduced an efficient and effective non-convex RPCA (eRPCA) that employed a faster rank estimation strategy and replaced Singular Value Decomposition (SVD) with Column-Row Decomposition, reducing the complexity of updating the low-rank matrix component from $O(rmnt)$ to $O(r^2mn)$ for $m \times n \times t$ input channel matrix with target rank $r$, While these methods achieved excellent background recovery and faster low-rank updates, they typically process color videos channel-by-channel, neglecting the structural correlation among the three channels.

It is noted that the method above generally performs well on two-dimensional matrices, failing to capture the inherent correlations across multiple dimensions. Tensor-based RPCA methods are powerful tools for multiple dimensions and have been widely studied in video analysis tasks such as moving target detection [23] and background recovery [12], [24]-[27]. However, the tensor SVD and triple decomposition are restricted to third-order tensors, while color videos are inherently in fourth order [28]. Recently, another promising direction has emerged with quaternion-based RPCA to process color image [29]-[31], which represents the color channels using the imaginary components of quaternions, and applies Quaternion SVD (QSVD) along with a complex adjoint formulation [32] to jointly process them, naturally preserving inter-channel correlations. However, performing QSVD on color videos incurs high computational cost. Moreover, while the largest singular values of a quaternion matrix contain the major color image information, e.g., color information and structural details [33], an ideal quaternion rank-1 matrix does not lead to rank-1 in each individual color channel, because pure quaternion matrices remain non-zero after operations.

In this paper, we address the initial question by introducing a universal Quaternion RPCA+ (uQRPCA+) method. Inspired by [34], which constrains the low-rank optimization process to a fixed-rank matrix manifold, this approach is particularly well-suited for background recovery, where ideal low-rank structures are often assumed. Motivated by this, we extend the method to the quaternion domain and propose a bidirectional quaternion weighting scheme to achieve more accurate singular value estimation, reducing the computational complexity of QSVD from $O(\min(m,n) \cdot mn)$ to $O(1)$. However, enforcing a strict rank-1 constraint in real-world scenarios often introduces noise that contaminates the target region, especially in dynamic background. Inspired by [19], which assumes that the dynamic background is generally sparser than the moving target, we incorporate an $l_1$-constrained SSB model with a novel bidirectional weighting mechanism. This mechanism adaptively adjusts the sparsity constraint during the iterative process, either strengthening or weakening, to recover finer details, albeit at the expense of increased noise. To address this issue, we incorporate isotropic 2D total variation (TV) regularization into the iterative framework, inspired by [35], to enhance denoising performance. This design achieves a balance between noise suppression and detail preservation, without compromising background recovery performance. Moreover, to maintain the ideal rank-1 for color channels, we further refine the recovered background using a Color Rank-1 Batch (CR1B) method. Remarkably, the method is highly efficient and can be implemented with just a few lines of MATLAB code.

We evaluate the moving target detection performance on the ChangeDetection.Net2014 (CDNet2014) dataset [36] and validate its background recovery capabilities on SceneBackgroundModeling.net (SBMnet) [37] and Scene Background Initialization (SBI) [38] datasets. The results demonstrate significant improvements over State Of The Art (SOTA) open-source methods, including the recently released deep learning approach Motion-aware Memory Network for Fast Video Salient Object Detection (MMN-VSO) [39] and Learning Temporal Distribution and Spatial Correlation (LTS) [40].

Our work presents several significant contributions to the field:

- A Fast Weighted Rank-1 Quaternion SVD (FWR1-QSVD) is proposed, which leverages quaternion Riemannian manifold optimization to more effectively approximate ideal singular values, reducing the computational complexity of QSVD from $O(\min(m,n) \cdot mn)$ to $O(1)$.
- We develop a uQRPCA framework along with an optimized algorithm that balances detailed target

extraction with minimal noise while enabling accurate color background recovery in real-world color videos.
- We further propose uQRPCA+, which employs the CR1B method to refine the background produced by uQRPCA, enforcing an ideal rank-1 structure within each color channel, and achieving SOTA results on both moving target detection and background recovery tasks. Furthermore, the uQRPCA+ method enables us to generate synthetic datasets, which can be used to train SOTA segmentation networks, thereby enhancing the generalization ability.

The rest of this paper is organized as follows: Section II presents the theoretical foundation of quaternion mathematics as applied in this study. Section III provides a detailed description of the uQRPCA+ framework. Section IV presents a comprehensive experimental evaluation. Section V summarizes and discusses potential avenues for future exploration.

## II. NOTIONS AND PRELIMINARIES

In this paper, scalar, vector, matrix, quaternion vector and quaternion matrix are represented by lowercase letters, bold lowercase letters, bold lowercase letters, bold uppercase letters, bold lowercase letters with an overhead dot, and bold uppercase letters with an overhead dot, respectively (e.g., $x$, $\boldsymbol{x}$, $\boldsymbol{X}$, $\dot{\boldsymbol{x}}$, and $\dot{\boldsymbol{X}}$).

**Definition 1 [41].** As an extension of the real space $\mathbb{R}$ and complex space $\mathbb{C}$, the quaternion space $\mathbb{H}$ was first introduced by Hamilton [42]. A quaternion matrix $\dot{\boldsymbol{N}}$, widely used in $m \times n \times 3$ color image processing, can be represented as:

$$\dot{\boldsymbol{N}} = \boldsymbol{N}_0 + \boldsymbol{N}_1 \boldsymbol{i} + \boldsymbol{N}_2 \boldsymbol{j} + \boldsymbol{N}_3 \boldsymbol{k} \in \mathbb{H}^{m \times n}, \quad (1)$$

where $\boldsymbol{N}_l \in \mathbb{R}^{m \times n}$ for $l = 0, 1, 2, 3$ are real matrices, and $\boldsymbol{i}, \boldsymbol{j}, \boldsymbol{k}$ are imaginary units satisfying the following relations:

$$\begin{aligned} \boldsymbol{ij} &= -\boldsymbol{ji} = \boldsymbol{k}, \\ \boldsymbol{jk} &= -\boldsymbol{kj} = \boldsymbol{i}, \\ \boldsymbol{ki} &= -\boldsymbol{ik} = \boldsymbol{j}, \\ \boldsymbol{i}^2 &= \boldsymbol{j}^2 = \boldsymbol{k}^2 = \boldsymbol{ijk} = -1. \end{aligned} \quad (2)$$

If $\boldsymbol{N}_0 = 0$, then $\dot{\boldsymbol{N}}$ is referred to as a pure quaternion matrix.

**Definition 2 [43] (Complex Adjoint Form).** Given a quaternion matrix $\dot{\boldsymbol{N}} \in \mathbb{H}^{m \times n}$, it can be equivalently expressed using a pair of complex matrices $\boldsymbol{N}_a, \boldsymbol{N}_b \in \mathbb{C}^{m \times n}$ as follows:

$$\dot{\boldsymbol{N}} = \boldsymbol{N}_a + \boldsymbol{N}_b \boldsymbol{j}, \quad (3)$$

where $\boldsymbol{N}_a = \boldsymbol{N}_0 + \boldsymbol{N}_1 \boldsymbol{i}$ and $\boldsymbol{N}_b = \boldsymbol{N}_2 + \boldsymbol{N}_3 \boldsymbol{i}$. Accordingly, the complex adjoint representation of the quaternion matrix $\dot{\boldsymbol{N}}$, is given by $\boldsymbol{C} \in \mathbb{C}^{2m \times 2n}$, and is defined as:

$$\boldsymbol{C} = \begin{bmatrix} \boldsymbol{N}_a & \boldsymbol{N}_b \\ -\boldsymbol{N}_b^\triangleleft & \boldsymbol{N}_a^\triangleleft \end{bmatrix} \in \mathbb{C}^{2m \times 2n}, \quad (4)$$

where $\triangleleft$ denotes the conjugation of a complex matrix.

**Definition 3 [30] (Quaternion Frobenius Norm).** For a quaternion matrix $\dot{\boldsymbol{N}} \in \mathbb{H}^{m \times n}$, the Frobenius norm is defined as:

$$\|\dot{\boldsymbol{N}}\|_F = \left( \sum_i \sum_j |\dot{\boldsymbol{n}}_{i,j}|^2 \right)^{\frac{1}{2}}, \quad (5)$$

where the element at position $(i,j)$ is $\dot{\boldsymbol{n}}_{i,j} = n_{i,j,0} + n_{i,j,1}\boldsymbol{i} + n_{i,j,2}\boldsymbol{j} + n_{i,j,3}\boldsymbol{k}$, with $n_{i,j,l} \in \mathbb{R}$ for $1 \le i \le m$, $1 \le j \le n$, and $l = 0, 1, 2, 3$, and its squared magnitude is defined by $|\dot{\boldsymbol{n}}_{i,j}|^2 = n_{i,j,0}^2 + n_{i,j,1}^2 + n_{i,j,2}^2 + n_{i,j,3}^2$.

**Definition 4 [30] (Quaternion Rank).** The rank of a quaternion matrix is defined as the number of non-zero singular values obtained from its QSVD.

**Theorem 1 [30] (QSVD).** Given a quaternion matrix $\dot{\boldsymbol{N}} \in \mathbb{H}^{m \times n}$ with rank $r$, there exist two unitary quaternion matrices $\dot{\boldsymbol{U}} \in \mathbb{H}^{m \times m}$ and $\dot{\boldsymbol{V}} \in \mathbb{H}^{n \times n}$, such that

$$\dot{\boldsymbol{N}} = \dot{\boldsymbol{U}} \boldsymbol{\Sigma} \dot{\boldsymbol{V}}^H, \quad (6)$$

where $\boldsymbol{\Sigma} \in \mathbb{R}^{m \times n}$ is a real diagonal matrix containing $r$ non-negative singular values in descending order, and $H$ denotes the quaternion conjugate transpose.

To enable simultaneous operations across all imaginary components, quaternion matrices are often converted to their complex adjoint form $\boldsymbol{C} \in \mathbb{C}^{2m \times 2m}$ for efficient computation. Specifically, the QSVD of a quaternion matrix $\dot{\boldsymbol{N}}$ can be reconstructed via the following formulation:

$$\begin{aligned} \dot{\boldsymbol{U}} &= c(\boldsymbol{U}_1) + c(-\boldsymbol{U}_2^\triangleleft)\boldsymbol{j}, \\ \boldsymbol{\Sigma} &= r\left(c(\boldsymbol{\Sigma}^\xi)\right), \\ \dot{\boldsymbol{V}} &= c(\boldsymbol{V}_1) + c(-\boldsymbol{V}_2^\triangleleft)\boldsymbol{j}, \end{aligned} \quad (7)$$

where $\boldsymbol{C} = \boldsymbol{U}_c \boldsymbol{\Sigma}^\xi \boldsymbol{V}_c^\dagger$, which is obtained via the complex singular SVD with $\boldsymbol{U}_c = [\boldsymbol{U}_1; \boldsymbol{U}_2] \in \mathbb{C}^{2m \times 2m}$, $\boldsymbol{V}_c = [\boldsymbol{V}_1; \boldsymbol{V}_2] \in \mathbb{C}^{2n \times 2n}$, and $\dagger$ denoting the conjugate transpose of a complex matrix. The $r(\cdot)$ and $c(\cdot)$ are used to extract the odd rows and odd columns of a complex matrix, respectively.

**Definition 5 [30] (Quaternion Nuclear Norm, QNN).** The quaternion nuclear norm $\|\dot{\boldsymbol{L}}\|_\circledcirc$ is defined as the sum of all non-zero singular values of $\dot{\boldsymbol{L}}$, i.e.,

$$\|\dot{\boldsymbol{L}}\|_\circledcirc = \sum_l \sigma_l(\dot{\boldsymbol{L}}), \quad (8)$$

where $\sigma_l(\dot{\boldsymbol{L}})$ denotes the $l$-th singular value of the quaternion matrix $\dot{\boldsymbol{L}}$, obtained via QSVD, and the summation is taken over all non-zero singular values.

**Theorem 2 [30] (QNN Minimization, QNNM).** Given a quaternion matrix $\dot{\boldsymbol{D}} \in \mathbb{H}^{m \times n}$, the optimal solution $\dot{\boldsymbol{L}} \in \mathbb{H}^{m \times n}$ that best approximates $\dot{\boldsymbol{D}}$ under low-rank regularization is given by the following optimization problem:

$$\min_{\dot{L}} \|\dot{D} - \dot{L}\|_F^2 + \lambda \|\dot{L}\|_{\circledast}, \qquad (9)$$

where $\lambda$ is a positive regularization parameter that balances the trade-off between the fidelity term $\|\dot{D} - \dot{L}\|_F^2$ and the regularization term $\|\dot{L}\|_{\circledast}$.

The optimal solution can be efficiently computed using the Quaternion Singular Value Thresholding (QSVT) algorithm:

$$\dot{L} = \dot{U} \mathcal{S}_\lambda(\Sigma) \dot{V}^H, \qquad (10)$$

where $\dot{U}, \Sigma, \dot{V}$ are obtained via the QSVD of $\dot{D}$, and the soft-thresholding operator $\mathcal{S}_\lambda(\Sigma) \in \mathbb{R}^{m \times n}$ is denoted as $\mathcal{S}_\lambda(\Sigma) = \mathrm{diag}(\max\{\sigma_l(\dot{D}) - \lambda, 0\})$ with $1 \leq l \leq r$.

## III. METHOD

### A. The Proposal of QRPCA

Considering that existing RPCA-based methods are primarily designed for real-valued two-dimensional matrix data, they often suffer from performance degradation when applied to color video processing. To address this limitation, a quaternion-based RPCA framework has been proposed, which more effectively captures inter-channel correlations by representing RGB videos in the quaternion domain. Specifically, in our formulation, a color video consisting of $t$ frames of size $m \times n \times 3$ is represented by a quaternion matrix $\dot{D} \in \mathbb{H}^{mn \times t}$, where each column corresponds to a vectorized RGB frame encoded in quaternion form. The matrixis $\dot{D}$ then decomposed into a low-rank background component $\dot{L} \in \mathbb{H}^{mn \times t}$ and a sparse target component $\dot{S} \in \mathbb{H}^{mn \times t}$.

Essentially, the model leverages QSVD, which transforms the quaternion matrix $\dot{L}$ into its complex adjoint form and applies complex SVD to simultaneously process the color channels, after which all results are transformed back into the quaternion domain and update the singular values. Notably, the larger singular values of a quaternion matrix preserve key color and structural information in color images [33]. This pattern is even more pronounced in color videos, where the first singular value far exceeds the second (see Fig. 2), unlike in grayscale or channels formats. Motivated by this observation, we enforce a rank-1 constraint on the quaternion matrix $\dot{L}$. To better accommodate diverse scene characteristics, we first introduce a Bidirectionally Weighted QNN (BWQNN) as

**Definition 6 (BWQNN).** Given $\dot{L} \in \mathbb{H}^{mn \times t}$ with rank $r$, the BWQNN of a quaternion matrix $\dot{L}$ is defined as $\|\dot{L}\|_{W_L} = \sum_l \omega_l \sigma_l(\dot{L})$ with $1 \leq l \leq r$, and weight $\omega_l$ is computed by

$$\omega_l = \frac{C_1 \sigma_l(\dot{L})}{(\sigma_{r+1}(\dot{L}) + \sigma_l(\dot{L})) e^\varepsilon}, \qquad (11)$$

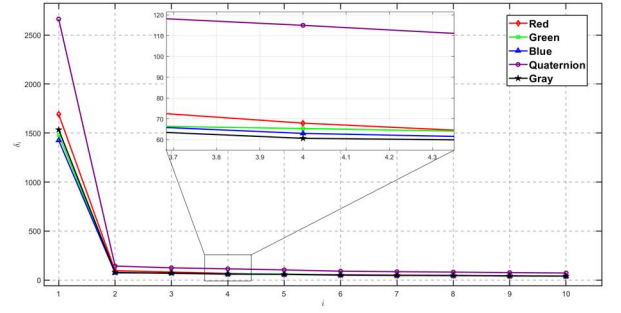

Fig. 2. The first 10 singular values are obtained via SVD of the IBMtest2 color video. The *Red*, *Green*, and *Blue* curves represent the singular values of the corresponding color channels, while the *Gray* curve corresponds to the grayscale version of the video. The curve labeled *Quaternion* represents the singular values derived from the quaternion-based representation of the color video.

where $C_1 \in \mathbb{R}$ is a tunable parameter that enables directional control over the singular value, and $\varepsilon = 10^{-4}$ is a small positive constant added for numerical robustness.

The resulting model is formulated as:

$$\min_{\dot{L}, \dot{S}} \|\dot{L}\|_{W_L} + \lambda \|\dot{S}\|_1 \\ s.t.\ \dot{D} = \dot{L} + \dot{S}, r = 1. \qquad (12)$$

Although this decomposition performs well under stationary background assumptions, its effectiveness diminishes significantly in real-world scenes with complex dynamics—such as waving curtains, water surfaces, lighting changes, or moving shadows. In such cases, a strict rank-1 constraint may lead to suboptimal target extraction. Inspired by [19], we decompose the $\dot{S}$ into both a target component $\dot{F} \in \mathbb{H}^{mn \times t}$ and a noise component $\dot{E} \in \mathbb{H}^{mn \times t}$, where both are assumed to be sparse but with different sparsity levels. To enhance sparsity and prevent target regions from being misclassified as noise, we apply an $l_1$-norm regularization to the SSB to enhance sparsity. Notably, our goal is to obtain an accurate target map. To reduce computational complexity during the iterative process, we sequentially and separately process the three imaginary components extracted from the quaternion matrices $\dot{S}, \dot{F}$ and $\dot{E}$.

To futher improve adaptability across datasets, we propose the adaptive Bidirectional Weighted $l_1$ Norm (BW $l_1$ N) defined as follows:

**Definition 7 (BW$l_1$N).** Given a sparse matrix $S \in \mathbb{R}^{mn \times t}$, the BW $l_1$ N is defined as $\|W_S \circ P_l(\dot{S})\|_1 = \sum_i \sum_j \omega_{i,j} |S_{i,j}|$ with $1 \leq i \leq mn$ and $1 \leq j \leq t$, $|S_{i,j}|$ denotes the magnitude of the sparse pixel at position $(i,j)$, and $\omega_{i,j}$ is the corresponding element of the weight matrix $W_S \in \mathbb{R}^{mn \times t}$, computed as:

$$\omega_{i,j} = C_2 \log(|S_{i,j}| + \varepsilon), \qquad (13)$$

where $\varepsilon$ is a small constant introduced to avoid numerical instability (i.e., to prevent undefined behavior when $|S_{i,j}|$=0),

and $C_2 > 0$ is a tunable parameter that controls the degree to which sparsity is suppressed or enhanced.

Its logarithmic weighting makes $\omega_{i,j}$ negative for small $|S_{i,j}|$, weakening sparsity, and positive for large $|S_{i,j}|$, strengthening sparsity. This enables automatic adjustment of sparsity during optimization to better handle complex scenes.

We perform denoising on the SSB using an image-domain denoising strategy. This is based on the observation that the 2-pass method employs an operator $P_l$ to reshapes the $l$-th SSB into a 2D matrix for iterative updates. Inspired by [35], which applies TV regularization to remove image noise during the iteration, we adopt 2D TV regularization in our framework. The incorporate TV regularization term $||\dot{F}||_{TV}$ is applied to each imaginary component of $\dot{F}$, and is defined as:

$$\|F\|_{TV} = \sum_i \sum_j \sqrt{(f_{i+1,j} - f_{i,j})^2 + (f_{i,j+1} - f_{i,j})^2}, \quad (14)$$

where $f_{i,j}$ denotes the pixel value at position $(i,j)$ in the reshaped 2D frame with $1 \le i \le m$ and $1 \le j \le n$.

The balances above get detailed target extraction with minimal noise while enabling accurate color background recovery in real-world color videos. The final formulation of our uQRPCA is as follows:

$$\min_{\dot{L},\dot{S},\dot{F},\dot{E}} \|\dot{L}\|_{W_L} + \sum_l \lambda_l \|W_S \circ P_l(\dot{S})\|_1 \\ + \rho_1 \|\dot{E}\|_1 + \rho_2 \|\dot{F}\|_{TV} \quad (15) \\ s.t. \dot{D} = \dot{L} + \dot{S}, \dot{S} = \dot{F} + \dot{E}, r = 1,$$

where $\rho_1$, $\rho_2$ and $\lambda_l$ are regularization coefficients controlling the trade-offs among different terms. Specifically $\lambda_l$ is defined as $\lambda_l = 0.1\overline{SM}_{\min}/\left(\overline{SM}_l\sqrt{\max(m,n)}\right)$, with $\overline{SM}_l$ denotes the motion saliency of the $l$-th block and $\overline{SM}_{\min}$ is the minimum saliency value across all blocks. This setting of $\lambda_l$ follows the same configuration as in the 2-pass method.

### B. The Optimization of QRPCA

We employ an Alternating Direction Method of Multipliers (ADMM) [43] to solve the proposed model (15). Specifically, this model can be optimized by minimizing the corresponding augmented Lagrangian formulation, given as follows:

$$\mathcal{L}(\dot{L},\dot{S},\dot{E},\dot{F},\dot{X},\dot{Y}) = \|\dot{L}\|_{W_L} \\ + \sum_l \lambda_l \|W_S \circ P_l(\dot{S})\|_1 + \rho_1 \|\dot{E}\|_1 + \rho_2 \|\dot{F}\|_{TV} \\ + \frac{\mu}{2}\|\dot{D} - \dot{L} - \dot{S}\|_F^2 + \frac{\mu}{2}\|\dot{S} - \dot{E} - \dot{F}\|_F^2 \quad (16) \\ + \langle \dot{X}, \dot{D} - \dot{L} - \dot{S}\rangle + \langle \dot{Y}, \dot{S} - \dot{E} - \dot{F}\rangle \\ s.t. r = 1,$$

where $\langle \dot{X}, \dot{D} - \dot{L} - \dot{S}\rangle = \dot{X}^H(\dot{D} - \dot{L} - \dot{S})$ denotes the quaternion matrix inner product [32], and $\mu > 0$ is the penalty operator. The variables $\dot{X}$ and $\dot{Y} \in \mathbb{H}^{mn \times t}$ are Lagrange Multipliers.

At each iteration $k$, the subproblems are solved by fixing all variables except one and updating them via minimization of the augmented Lagrangian:

1) **Update of $\dot{S}$:** To update $\dot{S}$, we fix all other variables and solve:

$$\dot{S}^{k+1} = \arg\min_{\dot{S}} \mathcal{L}(\dot{L}^k, \dot{S}, \dot{E}^k, \dot{F}^k, \dot{X}^k, \dot{Y}^k).$$

The model (16) reduces to three identical subproblems, each corresponding to one channel:

$$\min_{P_l(S)} \sum_l \frac{1}{2}\|P_l(Y_S) - P_l(S)\|_F^2 + \frac{\lambda_l}{\mu}\|W_S \circ P_l(S)\|_1. \quad (17)$$

Inspired by the soft thresholding operation, we introduce the novel Weighted Block Shrinkage (WBS) operator during the minimization:

$$WBS_{\epsilon_l}[P_l(Y_S)] \\ = \begin{cases} \frac{\|P_l(Y_S)\|_1 - \epsilon_l}{\|P_l(Y_S)\|_1} P_l(Y_S), & if \|P_l(Y_S)\|_1 > \epsilon_l, \\ 0, & otherwise, \end{cases} \quad (18)$$

where $\epsilon_l = \omega_{i,j}^k \lambda_l/\mu^k$ is the corresponding weighted threshold with weight $\omega_{i,j}^{k+1} = C_2 \log(|x_{i,j}^k| + \varepsilon)$.

After applying the WBS operator to all SSB, the updated $S$ is reconstructed by:

$$S^{k+1} = rP(WBS_{\epsilon_l}[P_l(Y_S)]), \quad (19)$$

where $Y_S = (D - L^k + E^k + F^k)/2 + (X^k - Y^k)/(2\mu^k)$, and $rP(\cdot)$ denotes the reconstruction of the vectorized matrix $\mathbb{R}^{mn \times t}$ after all SSB have been updated.

Finally, the quaternion sparse matrix $\dot{S}^{k+1}$ is updated by concatenating the three channels, i.e., $\dot{S}^{k+1} = cat(S^{k+1})$, where $cat(\cdot)$ denotes the reconstruction of a pure quaternion from the separately optimized RGB channels.

2) **Update of $\dot{L}$:** To update $\dot{L}$, we solve the following subproblem with all other variables fixed:

$$\dot{L}^{k+1} = \arg\min_{\dot{L}} \mathcal{L}(\dot{L}, \dot{S}^{k+1}, \dot{E}^k, \dot{F}^k, \dot{X}^k, \dot{Y}^k).$$

The problem reduces to:

$$\min_{\dot{L}} \frac{1}{2}\|\dot{Y}_L - \dot{L}\|_F^2 + \frac{1}{\mu}\|\dot{L}\|_{W_L} \quad (20) \\ s.t. r = 1,$$

where $\dot{Y}_L^k = \dot{D} - \dot{S}^{k+1} + 1/(\mu^k \dot{X}^k)$ and problem can be solved by QSVT.

Given the high computational complexity of QSVD in color videos processing and noting the inherent low-rank property of $\dot{Y}_L^k$, we follow the Riemannian optimization framework [34]. Specifically, we employ a projection operator $\pi(\cdot)$ that maps $\dot{Y}_L^k$ onto a rank-$r$ quaternion matrix lying on the smooth Riemannian manifold $M_r \in \mathbb{H}^{mn \times t}$, where:

$$M_r \in \mathbb{H}^{mn \times t} = \{\dot{L}^k \in \mathbb{H}^{mn \times t} : rank(\dot{L}^k) = r\}. \quad (21)$$

Its tangent space $\mathcal{T}$ at $\dot{L}^k = \dot{U}^k \Sigma^k \dot{V}^{k^H} \in \mathbb{H}^{mn \times t}$ is given by

$$\mathcal{T} = \left\{ [\dot{U}, \dot{U}_\perp] \begin{bmatrix} \mathbb{H}^{r \times r} & \mathbb{H}^{r \times (t-r)} \\ \mathbb{H}^{(mn-r) \times r} & 0^{(mn-r) \times (t-r)} \end{bmatrix} [\dot{V}, \dot{V}_\perp]^H \right\}, (22)$$

where $\dot{U}_\perp$ and $\dot{V}_\perp$ denote the orthonormal complements of $\dot{U}$ and $\dot{V}$, respectively. The optimization is performed within this subspace to reduce the search space and improve computational efficiency. The orthogonal projection of $\dot{Y}_L^k$ onto the tangent space $\mathcal{T}$ becomes:

$$\pi(\dot{Y}_L^k) = \dot{U}^k \dot{U}^{k^H} \dot{Y}_L^k + \dot{Y}_L^k \dot{V}^k \dot{V}^{k^H} - \dot{U}^k \dot{U}^{k^H} \dot{Y}_L^k \dot{V}^k \dot{V}^{k^H}. \quad (23)$$

A Householder QR decomposition is employed to further simplify this formula and reduce dimensionality. Specifically, we compute:

$$\dot{Q}_1^k \dot{R}_1^k = \dot{Y}_L^{k^H} \dot{U}^k (\dot{I} - \dot{V}^k \dot{V}^{k^H}), \quad (24)$$
$$\dot{Q}_2^k \dot{R}_2^k = (\dot{I} - \dot{U}^k \dot{U}^{k^H}) \dot{Y}_L^k \dot{V}_L^k, \quad (25)$$

where $\dot{Q}_1^k$, $\dot{R}_1^k$, $\dot{Q}_2^k$, and $\dot{R}_2^k$ are obtained by quaternion QR decomposition, and $\dot{I}$ denotes the quaternion identity matrix. This yields a compact form of the projection in (23) as follows:

$$\pi(\dot{Y}_L^k) = [\dot{U}^k, \dot{Q}_2^k] \begin{bmatrix} \dot{U}^{k^H} \dot{Y}_L^k \dot{V}^k & \dot{R}_1^{k^H} \\ \dot{R}_2^k & 0 \end{bmatrix} \begin{bmatrix} \dot{V}^{k^H} \\ \dot{Q}_1^{k^H} \end{bmatrix},$$
$$= [\dot{U}^k, \dot{Q}_2^k] \dot{Q}^k \begin{bmatrix} \dot{V}^{k^H} \\ \dot{Q}_1^{k^H} \end{bmatrix}, \quad (26)$$

Therefore, the problem reduces to optimizing $\dot{Q}^k$ and the problem becomes:

$$\min_{\widetilde{\dot{Q}}} \frac{1}{2} \|\dot{Q} - \widetilde{\dot{Q}}\|_F^2 + \frac{1}{\mu} \|\widetilde{\dot{Q}}\|_{W_L} \quad (27)$$
$$s.t. \, rank(\widetilde{\dot{Q}}) = r = 1,$$

where $\widetilde{\dot{Q}}$ denotes the desired quaternion low-rank approximation.

This problem is solved via the Weighted QSVT approach. In this approach, we first compute the QSVD of the matrix $\dot{Q}^k = \dot{U}_q^k \Sigma^k \dot{V}_q^{k^H}$. Then, the next iterate of the quaternion matrix $\widetilde{\dot{Q}}^{k+1}$ is updated as follows:

$$\widetilde{\dot{Q}}^{k+1} = \dot{U}_q^k S_{\tau W_L}(\Sigma^k) \dot{V}_q^{k^H}, \quad (28)$$

where $S_{\tau W_L}(\Sigma^k) = \text{diag}(\max\{\sigma_l(\dot{Q}^k) - \omega_l/\mu^k, 0\})$ with $\omega_l = C_2 \sigma_l(\dot{Q}^k)/(\sigma_{l+1}(\dot{Q}^k) + \sigma_l(\dot{Q}^k))e^\varepsilon$. Finally, we map the result back to the original space using the following formulation:

$$\widetilde{\dot{L}}^{k+1} = \dot{U}^{k+1} S_{\tau W_L}(\Sigma^k) \dot{V}^{k+1^H}, \quad (29)$$

where $\Sigma^{k+1} = S_{\tau W_L}(\Sigma^k)$, $\dot{U}^{k+1} = [\dot{U}^k, \dot{Q}_2^k] \dot{U}_q^k(:, 1:r)$ and $\dot{V}^{k+1} = [\dot{V}^k, \dot{Q}_1^k] \dot{V}^{k+1}(:, 1:r)$.

After performing quaternion operations, the real part of the result often remains non-zero and lacks clear physical meaning. To address this, we extract the three quaternion channels using the operation $\mathcal{C}(\cdot)$, which forms a new pure quaternion matrix. Specifically:

$$\dot{L}^{k+1} = \mathcal{C}(\widetilde{\dot{L}}^{k+1}). \quad (30)$$

**Remark:** we use the number of FLoating-point OPerations (FLOP) as a quantification metric for time complexity analysis. We discuss the time complexity of Truncated QSVD for each step.

Suppose quaternion matrix $\dot{L} \in \mathbb{H}^{mn \times t}$ with complex adjoint matrix complex adjoint form $C \in \mathbb{C}^{2mn \times 2t}$. When $mn \gg t$, to compute the Truncated SVD of a complex matrix $C \in \mathbb{C}^{2mn \times 2t}$, i.e., $C = U_c \Sigma^\xi V_c^\dagger$, we begin by performing eigen-decomposition on we begin by performing eigen-decomposition on $C^\dagger C$. Specifically,

$$C^\dagger C = V_c \Sigma^{\xi^\dagger} U_c^\dagger U_c \Sigma^\xi V_c^\dagger = V_c \Sigma^{\xi^2} V_c^\dagger = V_c \Lambda V_c^\dagger, \quad (31)$$

where $C^\dagger C \in \mathbb{C}^{t \times t}$ takes $\mathcal{O}(mnt^2)$ FLOP and $V_c \Lambda V_c^\dagger$ takes $\mathcal{O}(t^3)$ FLOP.

Then taking the square root of each eigenvalue (to recover the singular values) costs $\mathcal{O}(t)$ and sorting with complexit $\mathcal{O}(t \log_2 t)$.

The left singular vectors are then computed via $U_c = CV_c/\Sigma^\xi$ with $\mathcal{O}(t(mnt + mn))$ FLOP.

In total, the time complexity of Truncated SVD is

$$\mathcal{O}(2mnt^2 + t^3 + t + t \log_2 t + mnt) = \mathcal{O}(mnt^2).$$

Similarly, when $t \gg mn$, the total time complexity becomes $\mathcal{O}(t(mn)^2)$. Notably, computing the SVD of the complex adjoint matrix is equivalent to performing the QSVD on the corresponding quaternion matrix, with the added benefit of reduced computational complexity [30]. Moreover, the numerical updates involved in QSVT do not affect the overall time complexity, and thus the time complexity of QSVD remains $\mathcal{O}(\min(mn, t) \cdot mnt)$.

Since $\dot{Q}^k$ is a $2r \times 2r$ matrix, the complexity of QSVD is reduced from the original $\mathcal{O}(\min(mn, t) \cdot mnt)$ to $\mathcal{O}(r^3)$, ie., $\mathcal{O}(1)$ when $r = 1$.

3) *Update of $\dot{E}$:* To update $\dot{E}$, we solve the following subproblem by fixing all other variables:

$$\dot{E}^{k+1} = \arg\min_{\dot{E}} \mathcal{L}(\dot{L}^{k+1}, \dot{S}^{k+1}, \dot{E}, \dot{F}^k, \dot{X}^k, \dot{Y}^k).$$

The solution by the soft-thresholding operator of each channel is

$$E^{k+1} = \mathcal{S}_{\rho_1/\mu^k}\left[S^{k+1} - F^k + \frac{1}{\mu^k}Y^k\right], \quad (32)$$

where $\mathcal{S}_{\rho_1/\mu^k}(P^k) = \text{sgn}(P^k)\max\{|P^k| - \rho_1/\mu^k, 0\}$ with $P^k = S^{k+1} - F^k + \frac{1}{\mu^k}Y^k$.

The final quaternion update is given by $\dot{E}^{k+1} = cat(E^{k+1})$.

4) *Update of $\dot{F}$:* To update $\dot{F}$ we solve:

$$\dot{F}^{k+1} = \arg\min_{\dot{F}} \mathcal{L}(\dot{L}^{k+1}, \dot{S}^{k+1}, \dot{E}^{k+1}, \dot{F}, \dot{X}^k, \dot{Y}^k).$$

For each channel, this reduces to the following TV-regularized subproblem:

$$\min_F \rho_2 \|F\|_{TV} + \frac{\mu}{2}\|F - M\|_F^2, \quad (33)$$

where $F = [f_1, \cdots f_t] \in \mathbb{R}^{mn \times t}$, $M = [m_1, \cdots m_t] \in \mathbb{R}^{mn \times t}$ and $M = S^{k+1} - E^{k+1} + \frac{Y^k}{\mu^k}$ with $1 \le l \le t$. Each frame $f_l \in \mathbb{R}^{mn \times 1}$ of $F$ and $m_l \in \mathbb{R}^{mn \times 1}$ of $M$ are reshaped to 2D matrices $F_l, M_l \in \mathbb{R}^{m \times n}$. The per-frame subproblem becomes:

$$\min_{F_l} \rho_2 \|F_l\|_{2D-TV} + \frac{\mu}{2}\|F_l - M_l\|_F^2. \quad (34)$$

The above optimization problem (33) is divided into $t$ subproblems, and each is addressed by the gradient projection method [44]. Then the solution is reshaped to vectors $f_l^{k+1}$, and the updated matrix is constructed by:

$$F^{k+1} = [f_1^{k+1}, \cdots f_t^{k+1}]. \quad (35)$$

The final quaternion form $\dot{F}^{k+1} = cat(F^{k+1})$.

5) *Update of $\dot{X}, \dot{Y}$:* The Lagrange multipliers are updated as follows:

$$\dot{X}^{k+1} = \dot{X}^k + \mu^k(\dot{D} - \dot{L}^{k+1} - \dot{S}^{k+1}), \quad (36)$$
$$\dot{Y}^{k+1} = \dot{Y}^k + \mu^k(\dot{S}^{k+1} - \dot{E}^{k+1} - \dot{F}^{k+1}). \quad (37)$$

The algorithm is terminated upon reaching the maximum number of iterations.

---

**Algorithm 1:** Our algorithm for uQRPCA+.

**Input:** $\rho_1, \rho_2 > 0$, and observation batch matrix $D \in \mathbb{R}^{mn \times t \times 3}$.

**Initialization:**
$\dot{L}^0 = \dot{S}^0 = \dot{E}^0 = \dot{F}^0 = \dot{X}^0 = \dot{Y}^0 = 0 \in \mathbb{H}^{mn \times t}$,
$\dot{D} = \dot{U}^0 \Sigma^0 \dot{V}^0$,
$\mu^0 > 0, \rho = 1.5$ and $k = 0$.

**while** not k == 20 **do**
   Update each chaneel of $\dot{S}^{k+1}$ via (17).
   Update $\dot{L}^{k+1}$ via (20).
   Update each chaneel of $\dot{E}^{k+1}$ via (32).
   Update each chaneel of $\dot{F}^{k+1}$ via (33).
   Update multipliers via (36-37).
   $\mu^{k+1} = \mu^k \rho; k = k + 1$.

**End**

Convert $\dot{L} \in \mathbb{H}^{mn \times t}$ into RGB matrix $L \in \mathbb{R}^{mn \times t \times 3}$ and demonstrate the corresponding MATLAB code for CR1B method as follows:

for c = 1:3
  for row = 1:m*n
    mostFrequentValue = mode($L$ (row, :, c));
    minIdx = find($L$ (row, :, c) == mostFrequentValue, 1);
    $L$ (row, :, c) = $L$ (row, minIdx, c);
  end
end

**return** Optimal solution $(L, \dot{S}, \dot{E}, \dot{F})$.

---

*C. The Propose of uQRPCA+*

Despite the quaternion low-rank matrix $\dot{L} \in \mathbb{H}^{mn \times t}$ achieving rank-1 after quaternion operations, its real part is often nonzero. However, in practice, pure quaternions are commonly adopted for representing color videos. As a result, the omission of the real part makes the ranks of the extracted three channels $L \in \mathbb{R}^{mn \times t}$ not ideal low rank.

To address this issue, we develop a novel post-processing method to ensure that each channel achieves an ideal color low-rank structure, which is defined as follows:

**Definition 8 (Ideal Color Low Rank, ICLR).** Considering a pure quaternion $\dot{L} \in \mathbb{R}^{mn \times t}$ used to represent a color video, we say that $\dot{L}$ satisfies the ICLR property if the following two conditions hold:

1. The three imaginary components extracted from the quaternion matri $\dot{L}$ are each rank-1.

2. The $t$ columns of each imaginary components in $L$ are identical, i.e., each channel matrix is column-replicated.

Based on the ICLR property, we propose a simple, yet effective post-processing method called CR1B, which selects the maximum value across all columns as the representative value for each row.

By integrating CR1B into the uQRPCA framework, we extend it to uQRPCA+, enabling more refined processing of the color background and further improving the accuracy of background reconstruction.

## IV. EXPERIMENTAL RESULTS

In this section, we first evaluate the proposed uQRPCA+ framework on two color video processing tasks: moving target detection and background recovery. We then perform ablation studies to validate the contribution of each module. Next, we analyze the computational time of the low-rank update. Finally, we demonstrate the capability to generate synthetic datasets for potential applications in deep learning training.

***Datasets:*** For moving target detection, we test on the following sequences from CDNet2014 [36]: *overpass, fall, boats, canoe, pedestrians, blizzard, skating*, and *snowfall*. For background recovery, we use the *511* and *Blurred* sequences from SBMnet [37], as well as *IBMtest2*, *CAVIAR1*, *HighwayI*, and *HighwayII* from SBI [38]. To optimize computational efficiency and reduce memory usage, we randomly select 100 frames from the CDNet2014 dataset and 90 frames from each of the other datasets, with all videos downsampled to 320×240 resolution. It is worth noting that the SBI dataset uses PNG format for both input and output frames, while CDNet and SBMnet sequences use JPG format.

***Comparison Method:*** We evaluate the performance of our proposed uQRPCA+ model by comparing it with several methods, including: 2-pass [16], TVRPCA [19], TNN [24], eRPCA [22], Robust PCA via Half-Quadratic Function regularization (RPCA-HQF) [45], Convex–Convex Singular Value Separation (CCSVS) [25], Slim transform-based tensor RPCA (SRPCA) [27], MMN-VSOD [39], and LTS [40]. Furthermore, GT denotes the Ground Truth, which refers to the target data used for quantitative evaluation.

***Experimental Settings:*** For our method, all experimental parameters are fixed empirically across all tests: $\rho_1 = 2/\sqrt{m \times n}$, $\rho_2 = 0.035\sqrt{m \times n}$ and 20-iteration stopping criterion. Other methods use default settings. All experiments are conducted using MATLAB R2022b on a machine equipped with 16 GB RAM.

***Evaluation Metrics:*** To objectively assess target detection performance, we use three standard evaluation indicators: Recall (R), Precision (P), and F-measure (F). Higher values of these metrics reflect better detection performance. To quantitatively evaluate the background recovery performance, we adopt several widely used metrics from [38], including Average Gray-level Error (AGE), Percentage of Error Pixels (pEPs), Percentage of Clustered Error Pixels (pCEPs), Peak Signal-to-Noise Ratio (PSNR), Multi-Scale Structural Similarity Index (MS-SSIM), and Color image Quality Measure (CQM). Lower values for the first three indicate better recovery, and the latter and vice versa.

### A. Motion Target Detection Task

In this subsection, we demonstrate through experiments that our proposed method outperforms other approaches in moving target detection across various complex datasets.

Fig. 4 illustrates that RPCA-based methods such as TNN, eRPCA, RPCA-HQF, CCSVS, and SRPCA are commonly applied to background recovery tasks. These methods often enforce a strong low-rank constraint to obtain high-quality backgrounds. However, this comes at the cost of losing accuracy in target detection, especially in dynamic background scenarios, where moving background components are frequently misclassified as moving targets. The 2-pass method also encounters trade-off problems. Lowering the binarization threshold reveals more target details but introduces additional sparse noise, as demonstrated in Fig. 3. Furthermore, the 2-pass method tends to misclassify significant background motion as SSB. Even the correctly detected SSB often contains residual noise. This issue is evident in the *fall* data, where shaking leaves are erroneously detected as SSB, and in the *overpass* data, where leaves surrounding people cause noisy detections. Although TVRPCA shows strong noise suppression capabilities, it suffers from hollow regions within targets in the *overpass* and *snowfall* data, indicating an imbalance between target detection and noise removal. Deep learning methods achieve high detection performance on some datasets but perform poorly on others, demonstrating limited generalization ability. Overall, our uQRPCA+ captures more target details while effectively smoothing noise, achieving a balanced trade-off between denoising and accurate target extraction. It consistently produces superior recovery results across various challenging tasks.

Table I shows that methods focusing on background recovery tend to achieve high R rates due to their emphasis on low-rank approximation, which allows detection of most target pixels but leads to many false positives, resulting in low P. In contrast, methods primarily designed for target detection generally achieve higher P owing to denoising mechanisms but fail to detect a large portion of target pixels. Deep learning methods can achieve high F on certain data, but their overall performance suffers from poor generalization. Our proposed method outperforms all compared approaches by achieving the highest average R, P, and F, quantitatively demonstrating its superior generalization capability and its ability to balance target detection with minimal false alarms.

### B. Background Recovery Task

In this subsection, we compare the background recovery performance with several other approaches.

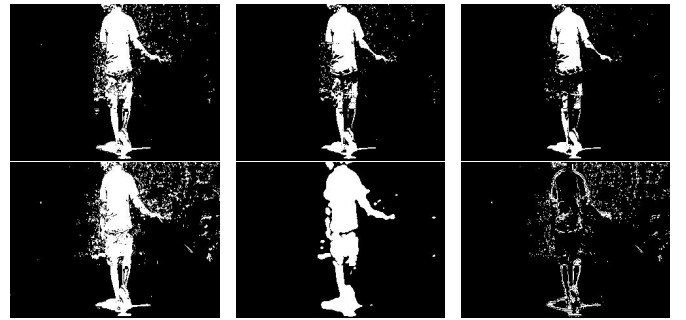

Fig. 3. The first row compares various thresholds of 2-pass for *overpass*: Left - 0.15, Middle - 0.2, and Right - 0.25. The second row displays the results of our uQRPCA with threshold of 0.11: Left - sparse foreground term, Middle - target term, Right - misclassification term.

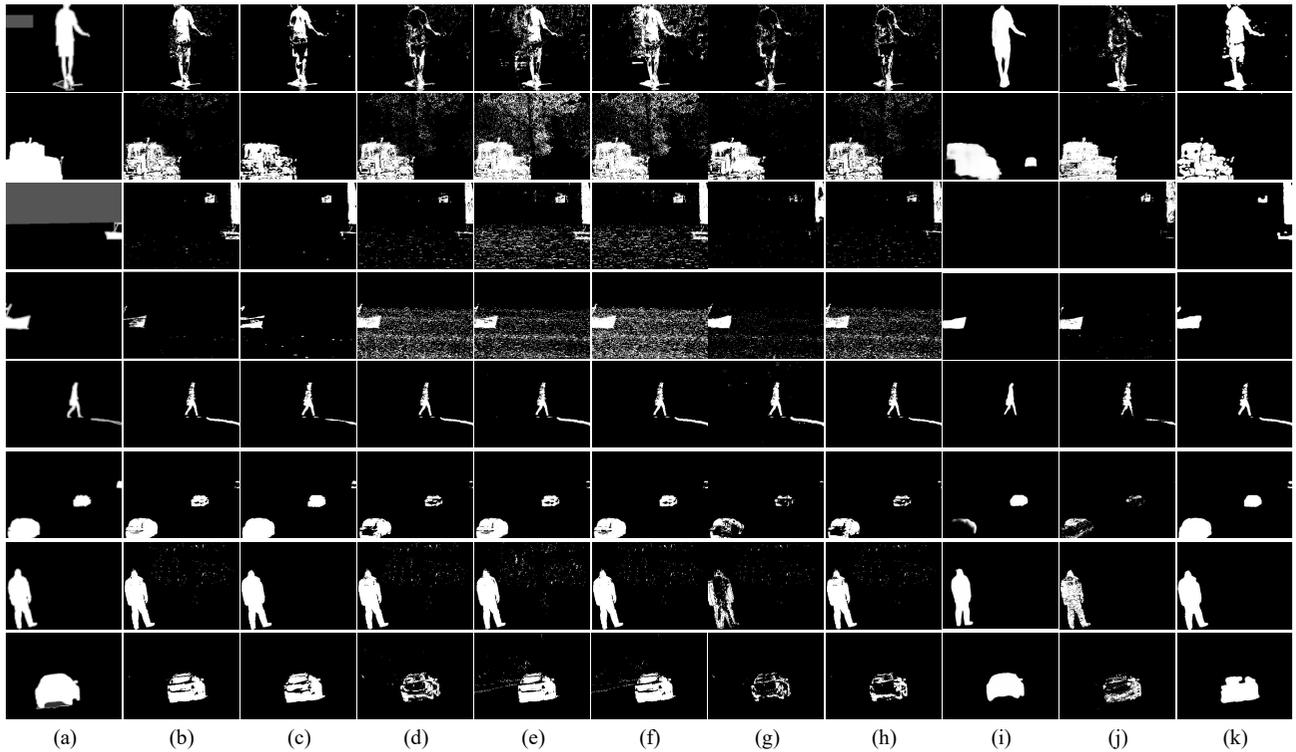

Fig. 4. Visual comparison of moving target detection results across various scenarios. Each row corresponds to a different video sequence, while each column represents the results of different methods. *overpass*, *fall*, *boats*, *canoe*, *pedestrians*, *blizzard*, *skating*, *snowFall*, going from top to bottom. From left to right: (a) GT (b) 2-pass (c) TVRPCA (d) TNN (e) eRPCA (f) RPCA-HQF (g) CCSVS (h) SRPCA (i) MMN-VSOD (j) LTS (k) uQRPCA+.

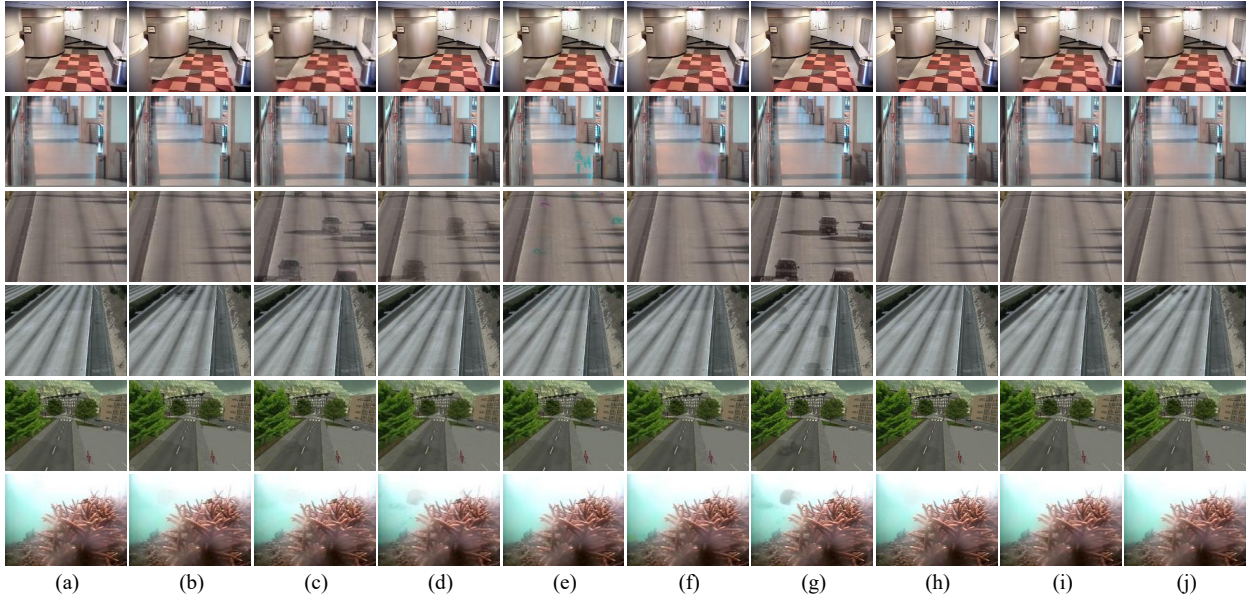

Fig. 5. Visual comparison of background recovery results across various scenarios. Each row corresponds to a different video sequence, while each column represents the results of different methods. *IBMtest2*, *CAVIAR1*, *HighwayI*, *HighwayII*, *511*, *Blurred*, going from top to bottom. From left to right: (a) GT (b) 2-pass (c) TVRPCA (d) TNN (e) eRPCA (f) RPCA-HQF (g) CCSVS (h) SRPCA (i) uQRPCA (j) uQRPCA+.

As illustrated in Fig. 5, target detection methods such as 2-pass, TNN, and TVRPCA lack the ability to adaptively modify the singular values of the low-rank component to accommodate dynamic scene variations during iteration, leading to suboptimal background recovery on some data. Although eRPCA achieves fast background recovery, it suffers from randomness in the recovered results. RPCA-HQF and CCSVS also struggle to produce clean backgrounds, whereas SRPCA, uQRPCA, and its extension uQRPCA+ consistently demonstrate superior background recovery quality.

Fig. 6 illustrates that most methods exhibit significant

TABLE I
COMPARISON OF MOVING TARGET DETECTION PERFORMANCE ON THE SEQUENCES SHOWN IN FIG. 7

| Evaluate | Method | 2-pass | TVRPCA | TNN | eRPCA | RPCA-HQF | CCSVS | SRPCA | MMN-VSOD | LTS | uQRPCA+ |
|---|---|---|---|---|---|---|---|---|---|---|---|
| R | overpass | 0.7928 | 0.7617 | 0.4428 | 0.7765 | 0.8944 | 0.4003 | 0.3514 | **0.9824** | 0.6541 | 0.9535 |
| | fall | 0.8394 | 0.7357 | 0.8259 | 0.9237 | 0.9014 | 0.8074 | 0.8009 | 0.4169 | **0.9353** | 0.8845 |
| | boats | 0.5570 | 0.1028 | 0.0963 | 0.5692 | 0.5642 | 0.0430 | 0.0926 | 0.0102 | 0.0256 | **0.6365** |
| | canoe | 0.4445 | 0.6327 | 0.9533 | 0.9040 | **0.9751** | 0.9553 | 0.9309 | 0.7541 | 0.9477 | 0.9504 |
| | pedestrians | 0.9486 | 0.8928 | 0.9412 | 0.9464 | 0.9480 | **0.9813** | 0.9388 | 0.9171 | 0.9211 | 0.9325 |
| | blizzard | 0.7866 | 0.9038 | 0.4151 | 0.7398 | 0.7543 | 0.3994 | 0.4525 | 0.7111 | 0.1039 | **0.9259** |
| | skating | 0.9930 | **0.9940** | 0.9873 | 0.9939 | 0.9922 | 0.4742 | 0.9889 | 0.9763 | 0.8561 | 0.9927 |
| | snowFall | 0.8549 | 0.8127 | 0.3843 | 0.8752 | 0.8418 | 0.1790 | 0.3049 | 0.8152 | 0.4150 | **0.9101** |
| | **Average** | 0.7771 | 0.7295 | 0.6308 | 0.8411 | 0.8589 | 0.5300 | 0.6076 | 0.6979 | 0.6074 | **0.8983** |
| P | overpass | 0.9807 | 0.9488 | 0.9005 | 0.7671 | 0.6948 | 0.9207 | 0.8767 | **0.9859** | 0.9495 | 0.9792 |
| | fall | 0.7359 | 0.9367 | 0.7646 | 0.2794 | 0.3461 | 0.7928 | 0.5898 | 0.6821 | 0.9699 | **0.9752** |
| | boats | 0.1901 | 0.4109 | 0.2828 | 0.0220 | 0.0380 | 0.1475 | 0.0750 | **1.0000** | 0.6235 | 0.9666 |
| | canoe | 0.9269 | 0.7033 | 0.3814 | 0.0820 | 0.0626 | 0.4939 | 0.1068 | **0.9971** | 0.9297 | 0.9677 |
| | pedestrians | 0.9436 | 0.9272 | 0.9631 | 0.9321 | 0.9433 | 0.8767 | 0.9727 | 0.9660 | 0.9541 | **0.9706** |
| | blizzard | 0.9998 | 0.9901 | 0.9998 | 0.9894 | **0.9999** | 0.9890 | 1.0000 | 0.9962 | 0.9932 | 0.9722 |
| | skating | 0.8290 | 0.9876 | 0.8850 | 0.2209 | 0.8158 | 0.4978 | 0.8826 | 0.3839 | 0.9961 | **0.9964** |
| | snowFall | 0.9711 | 0.9323 | 0.8965 | 0.9522 | 0.9650 | 0.9313 | **0.9848** | 0.8356 | 0.9786 | 0.9554 |
| | **Average** | 0.8221 | 0.8546 | 0.7592 | 0.5306 | 0.6082 | 0.7062 | 0.6861 | 0.8559 | 0.9243 | **0.9729** |
| F | overpass | 0.8768 | 0.8451 | 0.5937 | 0.7718 | 0.7821 | 0.2557 | 0.5018 | **0.9842** | 0.7746 | 0.9662 |
| | fall | 0.7842 | 0.8242 | 0.7941 | 0.4290 | 0.5002 | 0.8001 | 0.6794 | 0.5175 | **0.9522** | 0.9276 |
| | boats | 0.2834 | 0.1645 | 0.1438 | 0.0423 | 0.0712 | 0.0666 | 0.0829 | 0.0202 | 0.0491 | **0.6365** |
| | canoe | 0.6009 | 0.6661 | 0.5449 | 0.1504 | 0.1176 | 0.6511 | 0.1916 | 0.8588 | 0.9386 | **0.9590** |
| | pedestrians | 0.9461 | 0.9097 | 0.9520 | 0.9392 | 0.9456 | 0.9260 | **0.9555** | 0.9409 | 0.9373 | 0.9512 |
| | blizzard | 0.8805 | 0.9450 | 0.5866 | 0.8466 | 0.8599 | 0.5690 | 0.6231 | 0.8299 | 0.1881 | **0.9485** |
| | skating | 0.9036 | 0.9908 | 0.9333 | 0.3615 | 0.8954 | 0.4858 | 0.9327 | 0.5511 | 0.9208 | **0.9945** |
| | snowFall | 0.9093 | 0.8684 | 0.5380 | 0.9121 | 0.8992 | 0.3002 | 0.4656 | 0.8252 | 0.5828 | **0.9322** |
| | **Average** | 0.7731 | 0.7767 | 0.6358 | 0.5566 | 0.6339 | 0.5068 | 0.5541 | 0.6910 | 0.6679 | **0.9145** |

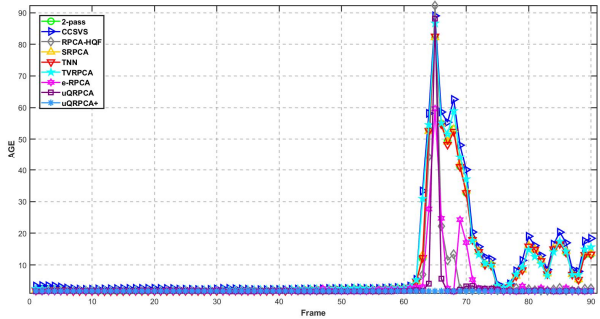

(a) The comparison of AQE on *Blurred* among different methods.

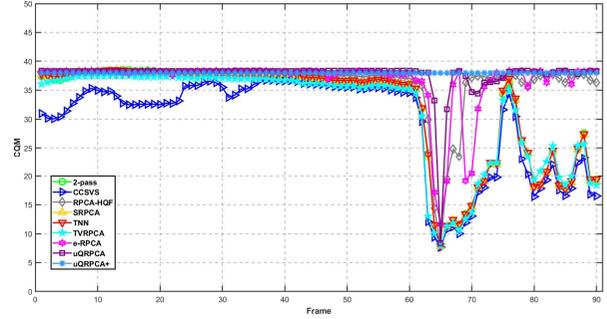

(b) The comparison of CQM on *Blurred* among different methods.

Fig. 6. Frame-wise comparison of different methods results on the *Blurred* sequence.

frame-to-frame variation in recovery quality, resulting in noticeable flickering effects. This issue is particularly pronounced in the *Blurred* sequence, where the background is completely occluded; the recovery quality for these frames is poor across all methods, as shown in Fig. 7. In contrast, results obtained by uQRPCA+ through the CR1B method maintain consistent quality across frames, thereby enhancing the overall video background recovery.

Table II quantitatively confirms that our method achieves SOTA results, especially in the *Blurred* data.

Additionally, we compare our method with the recent quaternion-based video processing technique Quaternion-based Dynamic Mode Decomposition (Q-DMD) [46]. Since its code is not publicly available and given the varying

> REPLACE THIS LINE WITH YOUR MANUSCRIPT ID NUMBER (DOUBLE-CLICK HERE TO EDIT) <

1TABLE II
COMPARISON OF BACKGROUND RECOVERY PERFORMANCE ON THE SEQUENCES SHOWN IN FIG. 7

| Evaluate | Method | 2-pass | TVRPCA | TNN | eRPCA | RPCA-HQF | CCSVS | SRPCA | uQRPCA | uQRPCA+ |
|---|---|---|---|---|---|---|---|---|---|---|
| AGE | *IBMtest2* | 3.3263 | 4.5079 | 4.1195 | 3.5219 | 3.1801 | 5.4917 | 3.6706 | 2.9269 | **1.9352** |
| | *CAVIAR1* | 1.4454 | 2.3785 | 3.7876 | 1.7973 | 2.2857 | 4.0227 | 3.3945 | 1.3999 | **1.2520** |
| | *HighwayI* | 2.1668 | 5.7663 | 4.7869 | 2.0661 | 1.9572 | 7.9971 | 2.1175 | 1.8255 | **1.7681** |
| | *HighwayII* | 3.3617 | 3.5433 | 3.7587 | 2.6081 | 2.4490 | 5.3449 | 2.5547 | 2.4232 | **2.3626** |
| | *511* | 3.9220 | 4.4357 | 4.1758 | 3.6442 | **3.5341** | 4.4965 | 3.8262 | 3.5677 | 3.5672 |
| | *Blurred* | 7.9917 | 8.6162 | 4.2450 | 3.2949 | 3.9950 | 9.5750 | 8.0063 | 2.8572 | **1.7306** |
| | **Average** | 3.7023 | 4.8747 | 4.1456 | 2.8221 | 2.9002 | 6.1547 | 3.9283 | 2.5001 | **2.1026** |
| pEPs | *IBMtest2* | 0.1504 | 0.3871 | 1.5973 | 0.3213 | 0.0408 | 3.9048 | 0.2560 | **0.0280** | 0.0339 |
| | *CAVIAR1* | 0.1427 | 0.8982 | 4.3135 | 0.7786 | 1.9505 | 4.3707 | 3.8454 | **0.1361** | 0.1393 |
| | *HighwayI* | 0.2627 | 8.2645 | 5.1777 | 0.3565 | 0.2665 | 12.9248 | 0.2075 | **0.2356** | 0.2461 |
| | *HighwayII* | 2.7274 | 0.9816 | 1.4154 | 0.8919 | 0.8529 | 5.5426 | 0.7839 | 0.7476 | 0.7500 |
| | *511* | 1.7189 | 2.5096 | 1.8469 | 0.9170 | **0.8547** | 2.5873 | 1.4129 | 0.9315 | 0.9271 |
| | *Blurred* | 8.7222 | 8.9836 | 3.4935 | 1.1774 | 2.9176 | 9.4980 | 8.7337 | 1.0859 | **0.0117** |
| | **Average** | 2.2874 | 3.6708 | 2.9741 | 0.7405 | 1.1472 | 6.4714 | 2.5399 | 0.5275 | **0.3514** |
| pCEPs | *IBMtest2* | 0.0031 | 0.0037 | 0.7850 | 0.0127 | **0.0000** | 2.2692 | 0.0316 | **0.0000** | **0.0000** |
| | *CAVIAR1* | 0.0625 | 0.5636 | 3.7340 | 0.2692 | 1.5355 | 3.7789 | 3.3063 | **0.0620** | 0.0625 |
| | *HighwayI* | 0.0393 | 5.7648 | 3.5366 | 0.0312 | 0.0364 | 10.0012 | 0.0269 | **0.0301** | 0.0313 |
| | *HighwayII* | 1.9190 | 0.1110 | 0.3340 | 0.0802 | 0.1289 | 2.8561 | 0.0841 | 0.0997 | 0.1003 |
| | *511* | 0.0406 | 0.0524 | 0.0260 | **0.0003** | 0.0000 | 0.0781 | 0.0165 | 0.0013 | 0.0012 |
| | *Blurred* | 7.8365 | 8.1175 | 2.5928 | 1.0523 | 2.6392 | 8.5523 | 7.8698 | 1.0595 | **0.0000** |
| | **Average** | 1.6502 | 2.4355 | 1.8347 | 0.2410 | 0.7233 | 4.5893 | 1.8892 | 0.2085 | **0.0326** |
| MSSSIM | *IBMtest2* | 0.9957 | 0.9876 | 0.9877 | 0.9925 | 0.9961 | 0.9674 | 0.9932 | 0.9963 | **0.9966** |
| | *CAVIAR1* | 0.9964 | 0.9824 | 0.9354 | 0.9811 | 0.9754 | 0.9295 | 0.9453 | **0.9969** | **0.9969** |
| | *HighwayI* | 0.9847 | 0.8561 | 0.9119 | 0.9885 | 0.9880 | 0.7861 | 0.9859 | 0.9900 | **0.9901** |
| | *HighwayII* | 0.9839 | 0.9826 | 0.9761 | 0.9808 | 0.9818 | 0.9333 | 0.9915 | **0.9923** | **0.9923** |
| | *511* | 0.9796 | 0.9733 | 0.9762 | **0.9869** | 0.9868 | 0.9702 | 0.9817 | 0.9860 | 0.9861 |
| | *Blurred* | 0.9418 | 0.9361 | 0.9741 | 0.9909 | 0.9905 | 0.9128 | 0.9423 | 0.9923 | **0.9953** |
| | **Average** | 0.9804 | 0.9530 | 0.9602 | 0.9868 | 0.9864 | 0.9166 | 0.9733 | 0.9923 | **0.9929** |
| PSNR | *IBMtest2* | 36.6973 | 33.0227 | 32.3942 | 34.7484 | 36.9898 | 29.1549 | 34.2359 | 37.8126 | **39.8685** |
| | *CAVIAR1* | 40.6296 | 35.3714 | 27.3879 | 37.2215 | 34.4192 | 26.7823 | 28.6635 | 40.8923 | **41.2698** |
| | *HighwayI* | 37.5487 | 28.3488 | 30.4108 | 37.6432 | 37.9464 | 24.6527 | 37.9016 | **38.5039** | 38.4257 |
| | *HighwayII* | 29.4201 | 32.3806 | 31.7134 | 32.4969 | 33.6616 | 28.6349 | 32.9077 | 34.2512 | **34.2936** |
| | *511* | 31.8802 | 30.8263 | 31.6370 | 33.0009 | **33.1229** | 30.7233 | 32.3151 | 33.0275 | 33.0291 |
| | *Blurred* | 32.8028 | 32.1286 | 32.8829 | 37.8485 | 37.3300 | 29.9884 | 32.8531 | 38.6029 | **39.3122** |
| | **Average** | 34.8298 | 32.0131 | 31.0710 | 35.4932 | 35.5783 | 28.3228 | 33.1462 | 37.1817 | **37.6998** |
| CQM | *IBMtest2* | 35.5111 | 31.8248 | 31.5231 | 33.4016 | 35.8952 | 28.3291 | 33.1719 | 36.8470 | **39.1412** |
| | *CAVIAR1* | 39.1512 | 34.0870 | 26.7360 | 36.5517 | 33.5763 | 26.1003 | 27.8853 | 39.4323 | **39.7718** |
| | *HighwayI* | 37.1825 | 28.2424 | 30.3081 | 36.8241 | 37.7158 | 24.8311 | 37.4993 | 38.2670 | **38.3549** |
| | *HighwayII* | 29.1982 | 31.9998 | 31.4053 | 31.6200 | 33.2860 | 28.4948 | 32.5575 | 33.8500 | **33.8886** |
| | *511* | 32.3513 | 31.3103 | 32.1163 | 33.3307 | 33.4207 | 31.2300 | 32.7427 | 33.3894 | **33.4270** |
| | *Blurred* | 32.0520 | 31.3807 | 32.2431 | 36.5147 | 36.3127 | 29.3848 | 32.1128 | 37.6624 | **38.3277** |
| | **Average** | 34.2411 | 31.4742 | 30.7220 | 34.7071 | 35.0345 | 28.0617 | 32.6616 | 36.5747 | **37.1519** |

recovery performance across different frames, we only used the *IBMtest2* dataset results reported in their paper, which also consists of 90 frames. As shown in Table IV, our method significantly outperforms Q-DMD.

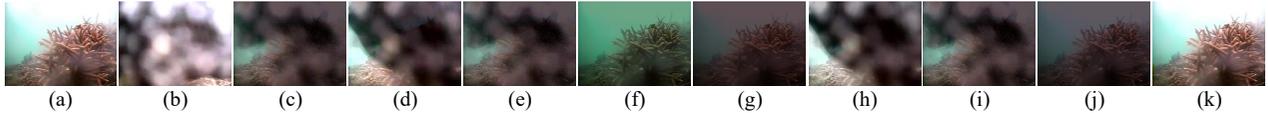

(a) (b) (c) (d) (e) (f) (g) (h) (i) (j) (k)

Fig. 7. Visual results of occluded frame recovery by different methods. From left to right: (a) GT (b) input (c) 2-pass (d) TVRPCA (e) TNN (f) eRPCA (g) RPCA-HQF (h) CCSVS (i) SRPCA (j) uQRPCA (k) uQRPCA+.

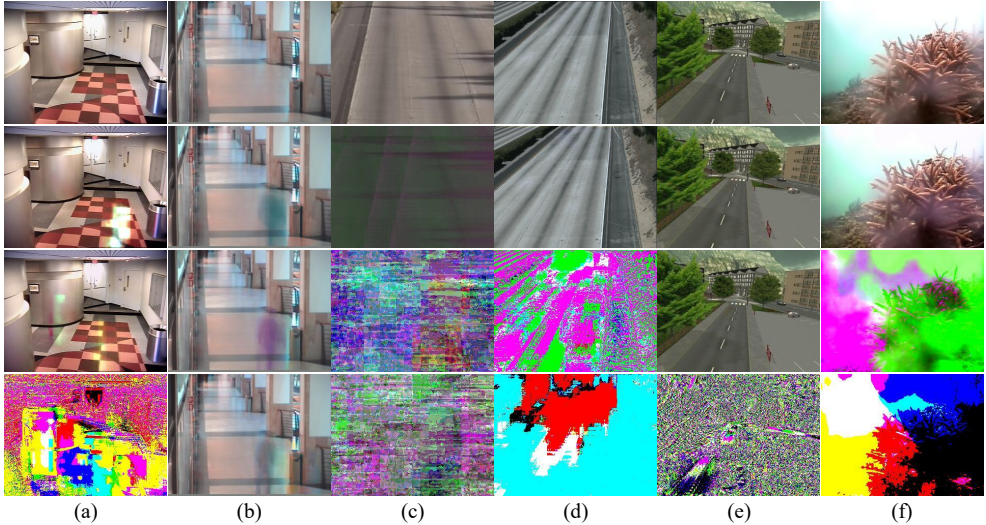

(a) (b) (c) (d) (e) (f)

Fig. 8. Visual results of our method, uQRPCA+, with different quaternion ranks. Each column corresponds to a different video sequence, while each row shows the results obtained using a different rank setting. Rank-1, Rank-2, Rank-3, Rank-4, going from top to bottom. From left to right: (a) *IBMtest2* (b) *CAVIAR1* (c) *HighwayI* (d) *HighwayII* (e) *511* (f) *Blurred*.

TABLE III
COMPARISON OF THE RESULTS UNDER DIFFERENT RANK SETTINGS ON THE SEQUENCES SHOWN IN FIG. 7. R DENOTES THE RED CHANNEL EXTRACTED FROM THE QUATERNION REPRESENTATION, WHILE G AND B CORRESPOND TO THE GREEN AND BLUE CHANNELS, RESPECTIVELY. KGDE, PROPOSED IN [22], IS A RANK ESTIMATION METHOD USED TO ESTIMATE THE IDEAL RANK OF EACH COLOR CHANNEL. UQRPCA+ DENOTES THE RANKS OF EACH COLOR CHANNEL OBTAINED VIA CR1B APPLIED TO THE QUATERNION RANK-1 RESULTS

| data | IBMtest2 | | | CAVIAR1 | | | HighwayI | | | HighwayII | | | 511 | | | Blurred | | |
|---|---|---|---|---|---|---|---|---|---|---|---|---|---|---|---|---|---|---|
| | R | G | B | R | G | B | R | G | B | R | G | B | R | G | B | R | G | B |
| $r=1$ | 4 | 4 | 4 | 4 | 4 | 4 | 4 | 4 | 4 | 4 | 4 | 4 | 4 | 4 | 4 | 4 | 4 | 4 |
| $r=2$ | 8 | 8 | 8 | 8 | 8 | 8 | 8 | 8 | 8 | 8 | 8 | 8 | 8 | 8 | 8 | 8 | 8 | 8 |
| $r=3$ | 12 | 12 | 12 | 12 | 12 | 12 | 12 | 12 | 12 | 9 | 9 | 9 | 12 | 12 | 12 | 12 | 12 | 12 |
| $r=4$ | 8 | 7 | 8 | 16 | 16 | 16 | 16 | 16 | 16 | 4 | 4 | 4 | 16 | 16 | 16 | 10 | 9 | 10 |
| KGDE | 1 | 1 | 1 | 1 | 1 | 1 | 1 | 1 | 1 | 1 | 1 | 1 | 1 | 1 | 1 | 1 | 1 | 1 |
| uQRPCA+ | 1 | 1 | 1 | 1 | 1 | 1 | 1 | 1 | 1 | 1 | 1 | 1 | 1 | 1 | 1 | 1 | 1 | 1 |

TABLE IV
COMPARISON OF UQCRPCA AND Q-DMD ON IBMTEST2

| Evaluate | Q-DMD | uQCRPCA |
|---|---|---|
| AGE | 4.8208 | **1.9352** |
| pEPs | 2.2513 | **0.0339** |
| pCEPs | 0.6133 | **0.0000** |
| MSSSIM | 0.9827 | **0.9966** |
| PSNR | 30.9653 | **39.8685** |
| CQM | 30.0359 | **39.1412** |

*C. Ablation Experimen*

In this section, we first investigate the impact of rank on the experimental results, followed by an empirical evaluation of whether the different components of our uQRPCA+ achieve a balanced performance.

Table III and Fig. 8 show that as the rank increases, the rank of the three-color channels extracted becomes uncertain, and the video background recovery significantly degrades. Even when the quaternion rank is set to 1, the rank of the three extracted color channels fails to reach the ideal rank of 1 as estimated by KGDE [22]. Overall, this indicates that the rank-1 constraint on the quaternion low-rank matrix is necessary, and that its results require further refinement.

To evaluate the contributions of different components within the framework, we remove one core improvement at a time and analyze its effect in the uQRPCA+ framework. Specifically, IMTV represents the removal of the target and noise terms, IMLOW denotes the removal of the bidirectional

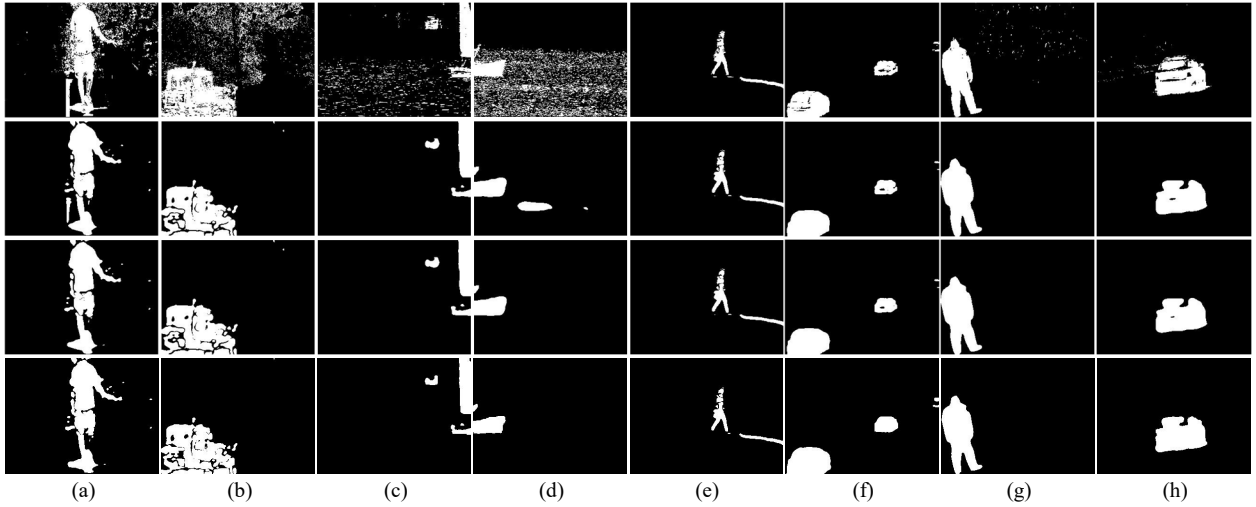

Fig. 9. Visual comparison of moving target detection results across various scenarios. Each column corresponds to a different video sequence, while each row represents the results of different methods. IMTV, IMLOW, IMSPARSE, uQRPCA+, going from top to bottom. From left to right: (a) *overpass* (b) *fall* (c) *boats* (d) *canoe* (e) *pedestrians* (f) *blizzard* (g) *skating* (h) *snowFall*.

TABLE V
COMPARISON OF ABLATION STUDY RESULTS ON THE SEQUENCES SHOWN IN FIG. 7

| Evaluate | Data | IMTV | IMLOW | IMSPARSE | uQRPCA+ |
|---|---|---|---|---|---|
| R | overpass | 0.9514 | **0.9544** | 0.9523 | 0.9535 |
|  | fall | **0.9074** | 0.8978 | 0.8880 | 0.8845 |
|  | boats | **0.7627** | 0.5514 | 0.5614 | 0.6365 |
|  | canoe | **0.9784** | 0.9643 | 0.9647 | 0.9504 |
|  | pedestrians | **0.9480** | 0.9466 | 0.9460 | 0.9325 |
|  | blizzard | 0.7973 | 0.8080 | 0.8153 | **0.9259** |
|  | skating | 0.9935 | 0.9966 | **0.9967** | 0.9927 |
|  | snowFall | 0.8753 | 0.8593 | 0.8640 | **0.9101** |
|  | **Average** | **0.9018** | 0.8723 | 0.87355 | 0.8983 |
| P | overpass | 0.8034 | 0.9599 | 0.9769 | **0.9792** |
|  | fall | 0.3481 | 0.9642 | 0.9654 | **0.9752** |
|  | boats | 0.0547 | 0.9233 | 0.9212 | **0.9666** |
|  | canoe | 0.0642 | 0.3829 | 0.9034 | **0.9677** |
|  | pedestrians | 0.9383 | 0.9546 | 0.9542 | **0.9706** |
|  | blizzard | 0.9993 | 0.9998 | **0.9999** | 0.9722 |
|  | skating | 0.8420 | 0.9587 | 0.9575 | **0.9964** |
|  | snowFall | 0.9617 | **0.9678** | 0.9657 | 0.9554 |
|  | **Average** | 0.6265 | 0.8889 | 0.9555 | **0.9729** |
| F | overpass | 0.8712 | 0.9572 | 0.9644 | **0.9662** |
|  | fall | 0.5032 | **0.9298** | 0.9251 | 0.9276 |
|  | boats | 0.1020 | 0.6904 | **0.6976** | 0.6365 |
|  | canoe | 0.1205 | 0.5482 | 0.9330 | **0.9590** |
|  | pedestrians | 0.9431 | 0.9506 | 0.9501 | **0.9512** |
|  | blizzard | 0.8872 | 0.8938 | 0.8982 | **0.9485** |
|  | skating | 0.9115 | 0.9773 | 0.9767 | **0.9945** |
|  | snowFall | 0.9165 | 0.9103 | 0.9120 | **0.9322** |
|  | **Average** | 0.6569 | 0.8572 | 0.9071 | **0.9145** |

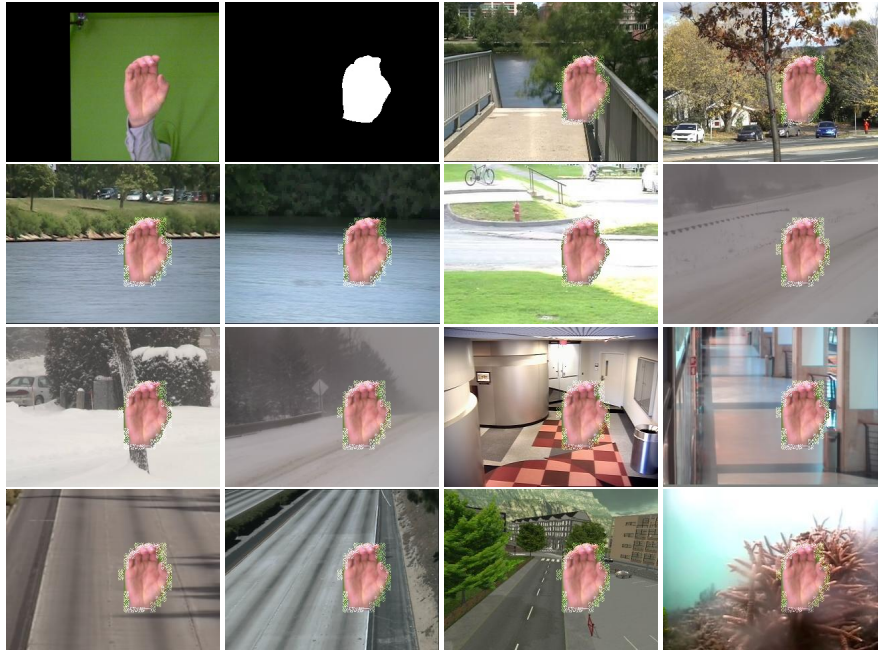

Fig. 10. Visual results on synthetic color images generated by our proposed method, uQRPCA+. The first row (from left to right) shows the original image and the ground-truth map. The remaining images illustrate newly synthesized hand mesh recovery data, obtained by compositing various recovered backgrounds with hand targets.

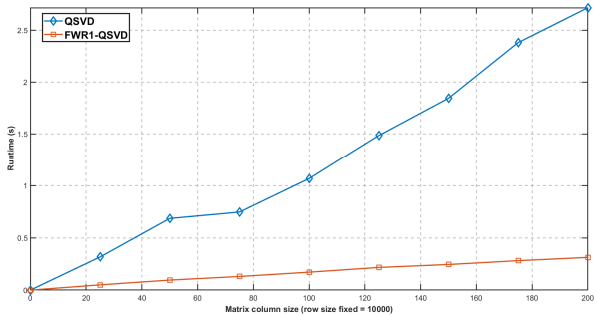

Fig. 11. Runtime comparison of the impact of different values of $n$ on low-rank update time using QSVD and FWR1-QSVD methods.

low-rank weight, and IMSPARSE indicates the absence of the bidirectional sparsity weight.

As shown in Fig. 9, the IMTV results clearly demonstrate that TV-based denoising is essential for effectively managing dynamic backgrounds. The IMLOW ablation study shows that removing the low-rank weighting leads to poor recovery of some background regions in the *canoe* sequence. The IMSPARSE results demonstrate that sparse weighting enables finer and dynamic adjustment of sparsity, leading to cleaner foreground extraction, as reflected in the *canoe* sequence.

As shown in Table V, the quantitative analysis reveals that IMTV achieves the highest R but the lowest P. IMSPARSE shows that the inclusion of sparse weighting further enhances both P and R, resulting in improved target detection performance. The results of IMLOW indicate that it significantly improves both P and R. Overall, our model achieves a well-balanced trade-off, accurately detecting targets while effectively removing noise and preserving a clean background.

*D. Low-Rank Update Time Comparison*

In this section, considering the significant time required for optical flow estimation, we use a synthetic dataset and focus on analyzing the impact of FWR1-QSVD that use Riemannian manifold optimization on the time complexity of the low-rank update.

We generate a matrix $C$, which is obtained by multiplying matrix $A \in \mathbb{R}^{m \times r}$ and matrix $B \in \mathbb{R}^{r \times n}$, where $r = 1$, $a_{ij} \in A \sim U(0,1)$ and $b_{ij} \in B \sim U(0,1)$. In our experiment, we aimed to simulate video data by fixing the row size at 10,000 and varying the column sizes across a list of 8 values: [25, 50, 75, 100, 125, 150, 175, 200], resulting in a total of 8 matrices denoted as $C$. Each matrix is two-dimensional, and we extend it into the quaternion $\dot{D}$ by setting the same $C$ on the $i, j, k$ vector coefficients. We employ the ADMM algorithm to solve the optimization problem, with the number of iterations fixed at 20. The average time per iteration is recorded and used for efficiency comparison.

Fig. 11 illustrates that as the column size increases, the time growth for QSVD significantly exceeds FWR1-QSVD. Since the matrix for QSVD is $2r \times 2r$, the complexity of QSVD is reduced from $O(mn^2)$ to $O(1)$. Although the theoretical complexity of quaternion QR decomposition is $O(mn^2)$, in practice, the FWR1-QSVD algorithm only performs QR decompositions on two thin matrices, $\mathbb{H}^{mn \times r}$ and $\mathbb{H}^{t \times r}$, respectively. As a result, the overall complexity of low-rank update is reduced to $O(\min(mnr, tr))$, ie., $O(\min(mn, t))$.

Our uQRPCA+ is applied to real-world datasets, the iteration time for low-rank updates decreases from an average of 3.2441s to 0.9091s on *overpass*, significantly reducing time consumption.

*E. Synthetic Data Experiments*

In this section, we evaluate the effectiveness of our method in generating synthetic data. As shown in Fig. 1, the target detection maps produced by our method closely match GT. Furthermore, by replacing the moving targets background in the *overpass* video with the recovered backgrounds, we can create more diverse new datasets. This demonstrates the feasibility of directly generating accurate target maps from in-the-wild datasets for supervised deep learning.

Fig. 10 further illustrates that the recovered backgrounds can be extended to other domains, such as hand mesh recovery, thereby enabling dataset augmentation in those fields.

## V. Conclusions and future directions

In this paper, we propose the uQRPCA+ framework, which enables accurate target detection and robust background recovery in in-the-wild color videos. Furthermore, this framework enables the generation of synthetic data. Future work will explore the impact of the synthesized data on deep learning methods.

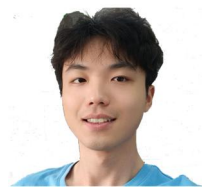

**Liyang Wang** is currently a junior undergraduate student at the School of Information Science and Engineering, Wuhan University of Science and Technology (WUST), Wuhan, China. His research interests include 3D/4D reconstruction and generation.

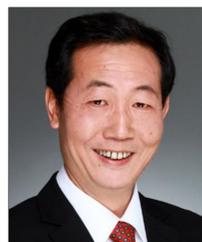

**Shiqian Wu** (Senior Member, IEEE) is a Professor at the School of Information Science and Engineering and the Deputy Director of the Institute of Robotics and Intelligent Systems (IRIS) at Wuhan University of Science and Technology (WUST), Wuhan, China. He is also the Director of the Hubei Province Key Laboratory of Intelligent Information Processing and Real-Time Industrial Systems, Wuhan, China. Dr. Wu received his B.Eng. and M.Eng. degrees from Huazhong University of Science and Technology (HUST), Wuhan, China, in 1985 and 1988, respectively, and his Ph.D. degree from Nanyang Technological University, Singapore, in 2001. Prior to joining WUST, he served as an Assistant Professor, Lecturer, and Associate Professor at HUST from 1988 to 1997. From 2000 to 2014, he was a Research Fellow and Research Scientist with the Agency for Science, Technology and Research (A-STAR), Singapore. Prof. Wu has co-authored two books and over 300 scientific publications, including book chapters and journal/conference papers. He received the BEST 10% PAPER award at ICIP 2015 and the recipient of the Best Paper Finalist award in ICIEA 2020. His research interests include Computer Vision, Pattern Recognition, Robotics, and artificial intelligence.

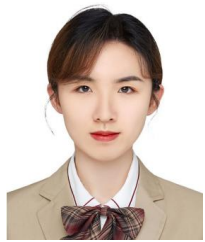

**Shun Fang** is currently a Ph.D. candidate at the School of Information Science and Engineering, Wuhan University of Science and Technology (WUST), Wuhan, China. She received her B.E. degree from WUST in 2019. Her research interests include robust principal component analysis and non-convex optimization.

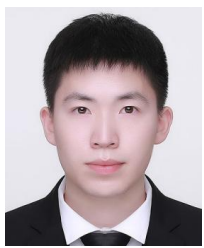

**Qile Zhu** is currently a Ph.D. candidate at the School of Information Science and Engineering, Wuhan University of Science and Technology (WUST), Wuhan, China. He received his B.E. degree from WUST in 2019. His research interests include tensor decomposition and high-dimensional signal processing.

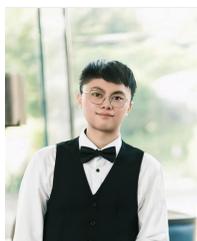

**Jiaxin Wu** graduated from the School of Science, Wuhan University of Science and Technology, China with a Bachelor of Science degree in 2017. He is currently a Ph.D. candidate in Institute of Robotics and Intelligent Systems, School of Information Science and Engineering, Wuhan University of Science and Technology, China. He has published several papers in Journal of Electronic Imaging, Signal, Image and Video Processing, IEEE Signal Processing Letters, IEEE Computer Graphics and Applications and other journals. He has also published several conference papers and won the second best paper award at the 15th IEEE Conference on Industrial Electronics and Applications (ICIEA 2020). His research interests include low-rank matrix decomposition, signal processing and machine learning, image processing and computer vision.

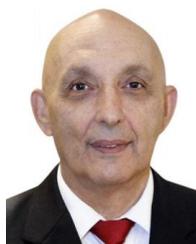

**Sos Agaian** (Fellow, IEEE) is a Distinguished Professor of Computer Science at the City University of New York (CUNY). He holds a Ph.D. in Mathematics and a Doctor of Engineering Science degree. His research spans a wide range of fields, including artificial intelligence, computational vision, cancer imaging, encryption and compression, image quality assessment, image editing, and multimedia systems. His work is driven by the goal of enabling computers to perceive, learn, and understand the world in ways similar to human cognition. Dr. Agaian has published over 800 scientific articles, authored 10 books, contributed 19 book chapters, and holds 56 patents and disclosures. He has supervised more than 45 Ph.D. students and has successfully commercialized many of his intellectual properties. He serves as an Associate Editor for leading journals such as *IEEE Transactions on Image Processing* and *IEEE Transactions on Cybernetics*. He is a Fellow of IS&T, SPIE, AAAS, IEEE, and AAIA, and a Member of the Academia Europaea. He is also a Distinguished Lecturer of the IEEE Systems, Man, and Cybernetics (SMC) Society. Dr. Agaian has delivered over 35 plenary and keynote speeches and more than 100 invited talks. He has co-founded and served as honorary or general co-chair and technical committee member for numerous international conferences in image processing and computer vision. He was the recipient of the Innovator of the Year Award (2014) and the Tech Flash Titans – Top Researcher Award from the *San Antonio Business Journal*. His research has been funded by prestigious agencies including AFRL, NSF, NIH, DARPA, the U.S. Army, and the Department of Transportation.